\documentclass[]{youtu} 

\usepackage{mathpazo}
\usepackage{fontawesome}
\usepackage{graphicx}
\usepackage[numbers]{natbib}
\usepackage{listings}
\usepackage{xcolor}
\usepackage{multirow}
\usepackage{subcaption}
\usepackage{fdsymbol}
\lstdefinestyle{yaml}{
  basicstyle=\ttfamily\footnotesize,
  columns=fullflexible,
  frame=single,
  backgroundcolor=\color{gray!10},
  breaklines=true,
  keywordstyle=\color{blue},
  commentstyle=\color{gray},
  stringstyle=\color{orange},
}

\title{
  \raisebox{-0.1\height}{\includegraphics[height=1.5em]{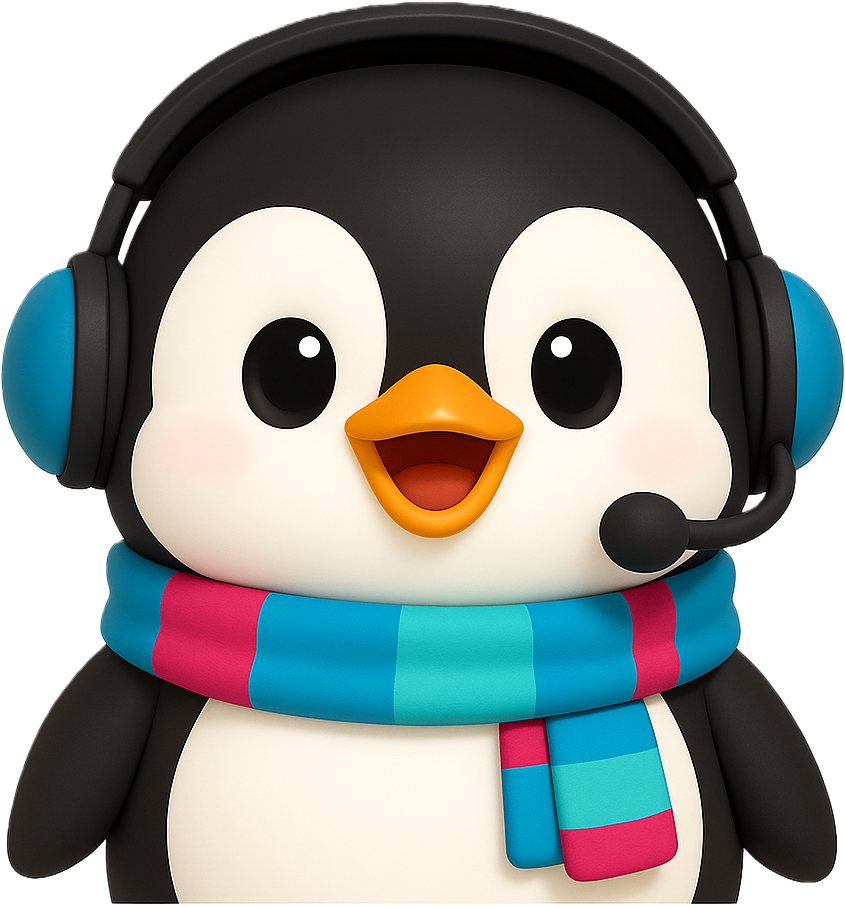}}
  Youtu-Agent: Scaling Agent Productivity with Automated Generation and Hybrid Policy Optimization
}
\author{Youtu-Agent Team$^*$}

\date{December 26, 2025}
\sourcecode{https://github.com/TencentCloudADP/youtu-agent}

\begin{document}

\abstract{
  Existing Large Language Model (LLM) agent frameworks face two significant challenges: high configuration costs and static capabilities. Building a high-quality agent often requires extensive manual effort in tool integration and prompt engineering, while deployed agents struggle to adapt to dynamic environments without expensive fine-tuning. To address these issues, we propose \textbf{Youtu-Agent}, a modular framework designed for the automated generation and continuous evolution of LLM agents. Youtu-Agent features a structured configuration system that decouples execution environments, toolkits, and context management, enabling flexible reuse and automated synthesis. We introduce two generation paradigms: a \textbf{Workflow} mode for standard tasks and a \textbf{Meta-Agent} mode for complex, non-standard requirements, capable of automatically generating tool code, prompts, and configurations. Furthermore, Youtu-Agent establishes a hybrid policy optimization system: (1) an \textbf{Agent Practice} module that enables agents to accumulate experience and improve performance through in-context optimization without parameter updates; and (2) an \textbf{Agent RL} module that integrates with distributed training frameworks to enable scalable and stable reinforcement learning of any Youtu-Agents in an end-to-end, large-scale manner.
  Experiments demonstrate that Youtu-Agent achieves state-of-the-art performance on WebWalkerQA (71.47\%) and GAIA (72.8\%) using open-weight models.
  Our automated generation pipeline achieves over 81\% tool synthesis success rate, while the Practice module improves performance on AIME 2024/2025 by +2.7\% and +5.4\% respectively.
  Moreover, our Agent RL training achieves 40\% speedup with steady performance improvement on 7B LLMs,
  enhancing coding/reasoning and searching capabilities respectively up to 35\% and 21\% on Maths and general/multi-hop QA benchmarks.
}
\maketitle

\renewcommand{\thefootnote}{*}
\footnotetext{Full author list in contributions.}
\renewcommand{\thefootnote}{\arabic{footnote}}

\vspace{-.1em}


\section{Introduction}

As Large Language Models (LLMs) evolve towards general purpose agents, building systems that autonomously solve complex real-world tasks has become a central pursuit in AI research. However, the current landscape of agent development is hindered by two primary bottlenecks.

\textbf{High Configuration Costs.} Developing a robust agent remains a labor-intensive process. On one hand, practitioners must select appropriate tools and craft effective prompts, which require significant domain expertise and iterative refinement. On the other hand, implementing custom tools demands substantial engineering effort, including writing Python functions, integrating external APIs, and handling edge cases. This ``craftsman'' approach creates a high barrier to entry and limits the scalability of agent deployment.

\textbf{Static Capabilities.} Once deployed, most existing agents remain static and struggle to adapt to new environments. Improving their performance typically requires one of two approaches: (1) manual optimization of prompts or few-shot examples, which is costly and offers no guarantee of improvement; or (2) Supervised Fine-Tuning (SFT) or Reinforcement Learning (RL), which face substantial challenges including data scarcity, high computational costs, and training instability issues such as ``entropy explosion'' in long-horizon tasks.

To overcome these challenges, we present \textbf{Youtu-Agent}, a comprehensive framework that bridges the gap between ``automated construction'' and ``continuous optimization.'' To address high configuration costs, Youtu-Agent adopts a layered, modular architecture that decouples \textbf{Environment} (execution context), \textbf{Toolkits} (atomic operations), and \textbf{Agent} (LLM-driven planning) into standardized components. This design enables a YAML-based structured configuration system that facilitates component reuse and serves as the foundation for automated generation. Building on this architecture, we provide two generation paradigms: the \textbf{Workflow} mode offers a deterministic pipeline for routine tasks, while the \textbf{Meta-Agent} mode deploys a higher-level architect agent to handle complex, dynamic requirements.

To address static capabilities, Youtu-Agent establishes a hybrid policy optimization system. For low-cost scenarios where manual prompt optimization proves unreliable, we introduce the \textbf{Agent Practice} module, which enables agents to accumulate experience and self-evolve during runtime. By performing parallel rollouts on dozens of samples and leveraging group relative experiential knowledge optimization, the agent builds a contextual memory that enhances future performance without gradient updates. For scenarios demanding significant and lasting performance improvement,
the \textbf{Agent RL} module integrates with existing RL training frameworks and bridging protocols to provide an end-to-end reinforcement learning recipe.
To ensure both stability and scalability during RL, our module tackles the concurrency bottleneck and entropy explosion/collapsing challenges respectively from the infrastructure and algorithm implementations.


Our contributions are as follows:

\begin{itemize}
  \item \textbf{Framework Efficiency.} Built entirely on open-source models and tools, Youtu-Agent achieves 71.47\% pass@1 on WebWalkerQA and 72.8\% pass@1 on GAIA (text-only subset), establishing a strong open-source baseline and validating the framework's general applicability.
  \item \textbf{Automated Agent Generation.} Through our two-paradigm generation mechanism, we demonstrate the potential of LLMs to automatically generate tools and agent configurations, significantly reducing the burden of manual configuration. The automated tool synthesis achieves over 81\% success rate.
  \item \textbf{Continuous Experience Learning.} Through the Agent Practice module, we enable low-cost continuous evolution of agents via in-context experience accumulation. This approach yields +2.7\% and +5.4\% absolute improvements for DeepSeek-V3.1-Terminus on AIME 2024 and AIME 2025, respectively.
  \item \textbf{Scalable and Stable Agent RL.} Through the Agent RL module, we address the scalability and stability challenges in large-scale reinforcement learning training. Our optimizations achieve a 40\% speedup in training iteration time and improve Qwen2.5-7B accuracy on AIME 2024 from 10\% to 45\%, demonstrating the framework's potential for building evolvable agents.
\end{itemize}

\section{Youtu-Agent Framework}

Youtu-Agent is designed with modularity and automation at its core. The framework comprises three hierarchical layers---Environment, Tools, and Agent---supported by a structured configuration system that enables both manual flexibility and automated synthesis. Beyond the execution architecture, the framework provides components for continuous optimization: the Agent Practice module enables low-cost experience-based improvement, while the Agent RL module supports end-to-end reinforcement learning. In this section, we describe each component in detail.

\subsection{System Architecture}

\subsubsection{Execution Framework}

The agent execution architecture is organized into three hierarchical layers. This layered design decouples concerns and enables flexible reuse of components across different agents.

\textbf{Environment Layer.} The Environment Layer provides the foundational execution context and low-level capabilities. It serves as the substrate upon which tools operate, exposing both state information (such as current page content or filesystem status) and primitive interfaces. Typical backends include browser instances for web navigation (e.g., Playwright-based browser environments), operating system shells for command execution, and sandboxed code execution environments (e.g., E2B). The environment abstraction allows the same tools and agents to operate across different backends with minimal modification.

\textbf{Tools Layer.} The Tools Layer encapsulates atomic and composite operations behind stable interfaces that the agent can invoke. Tools fall into three categories: (1) environment-related tools that wrap low-level environment APIs (e.g., clicking a DOM element, running a bash command, taking screenshots), (2) environment-independent utilities that perform standalone computations (e.g., mathematical operations, text processing, date/time handling), and (3) MCP tools that integrate external Model Context Protocol services for extended capabilities. This categorization allows tools to be composed, shared, and reused across different agents and environments.

\textbf{Agent Layer.} The Agent Layer houses the LLM-driven planner/executor that orchestrates task completion. The agent operates through a perceive--reason--act loop: it perceives environment state provided by the Environment Layer, reasons about the next steps using LLM capabilities, and selects appropriate tools to act. To handle long-horizon interactions that can overwhelm the model's context window, we incorporate a Context Manager module within this layer. The Context Manager maintains a compact working context by pruning stale or redundant information while preserving essential state. For example, in browser-based tasks, it removes obsolete HTML from prior navigation steps to control token cost without losing task-critical history.

\subsubsection{Configuration System}

A distinguishing feature of Youtu-Agent is its YAML-based structured configuration system. All components---environment specification, tool selection, agent instructions, and context management settings---are declared via human-readable YAML files. This standardized format serves dual purposes: it facilitates manual composition and sharing of agent configurations, and it provides the target schema for our automated generation mechanisms (Section~\ref{sec:autogen}). An example configuration is shown below:

\begin{lstlisting}[style=yaml]
agent:
  name: research_agent
  instructions: "You are a helpful research assistant..."
env:
  name: e2b
  config: {}
context_manager:
  name: base
  config: {}
toolkits:
  search:
    activated_tools: ["search", "web_qa"]
  python_executor:
    activated_tools: ["execute_python_code"]
\end{lstlisting}

This configuration specifies an agent with E2B sandbox environment, the default context manager, and two toolkits for web search and Python code execution. The declarative nature of YAML configurations enables rapid prototyping and systematic management of agent variants.

\subsubsection{Continuous Optimization Components}

Beyond execution, the framework emphasizes continuous agent improvement. Three supporting components enable this capability:

\begin{itemize}
  \item \textbf{Eval}: Evaluates agent performance on predefined benchmarks, providing standardized metrics for comparison and iteration.
  \item \textbf{Agent Practice}: Enables low-cost experience accumulation on small datasets without parameter updates (Section~\ref{sec:practice}).
  \item \textbf{Agent RL}: Supports end-to-end reinforcement learning training for scenarios requiring peak performance through parameter optimization (Section~\ref{sec:agentrl}).
\end{itemize}

These components transform Youtu-Agent from a static execution framework into an evolvable system capable of continuous self-improvement.

\subsection{Automated Generation Mechanism}
\label{sec:autogen}

Building upon the structured configuration system, Youtu-Agent provides automated mechanisms for generating complete agent configurations from high-level task descriptions. We introduce two generation paradigms tailored to different complexity levels and use cases, as illustrated in Figure~\ref{fig:autogen}.

\begin{figure}[htbp]
  \begin{center}
    \includegraphics[width=0.95\linewidth]{figs/fig2_autogen.png}
  \end{center}
  \caption{Automated generation mechanism. Left: user input describing the desired agent. Middle: two generation paradigms---Workflow mode (top) follows a deterministic four-stage pipeline, while Meta-Agent mode (bottom) deploys an architect agent with flexible tool access. Right: the generated agent configuration ready for deployment.}
  \label{fig:autogen}
\end{figure}

\subsubsection{Workflow Mode}

The Workflow mode follows a deterministic, four-stage pipeline suitable for well-defined, routine agent construction tasks:

\textbf{Stage 1: Intent Clarification and Decomposition.} The user's high-level task description is analyzed and decomposed into specific technical requirements. This stage identifies core objectives, necessary capabilities, and potential environmental constraints, producing a structured specification for subsequent stages.

\textbf{Stage 2: Tool Retrieval and Ad-hoc Tool Synthesis.} Based on the clarified requirements, the system searches the existing toolkit library for relevant tools. If required tools are missing, the \textbf{Ad-hoc Tool Synthesis} module automatically generates new Python tool implementations, complete with function signatures, docstrings, and unit tests. This synthesis process leverages LLMs to produce executable code that integrates seamlessly with the framework's tool interface.

\textbf{Stage 3: Prompt Engineering.} With the selected or synthesized tools identified, this stage generates optimized system instructions. The prompt engineering process considers task requirements, tool usage patterns, and desired agent behavior to produce effective instructions.

\textbf{Stage 4: Configuration Assembly.} The final stage compiles all components---environment specification, selected tools, generated prompts, and context management settings---into a complete YAML configuration file ready for deployment.

This workflow provides a standardized, reproducible pipeline for agent generation, enabling rapid development of agents for common task categories.

\subsubsection{Meta-Agent Mode}

For complex or ambiguous requirements that resist a fixed pipeline, the Meta-Agent mode deploys a higher-level Architect Agent that treats generation capabilities as callable tools. The meta-agent is equipped with: \texttt{search\_tool} (retrieve existing tools from the library), \texttt{create\_tool} (synthesize missing Python tools with signatures, docstrings, and unit tests), \texttt{ask\_user} (gather missing constraints or preferences through multi-turn dialogue), and \texttt{create\_agent\_config} (assemble the final YAML configuration).

With these capabilities, the meta-agent can dynamically plan the generation process: it conducts multi-turn clarification with \texttt{ask\_user} to elicit task scope and constraints, alternates between \texttt{search\_tool} and \texttt{create\_tool} depending on toolkit coverage, and finally integrates all components via \texttt{create\_agent\_config}. 

\subsection{Continuous Experience Learning}
\label{sec:practice}

How can we adapt an agent to specific tasks without expensive parameter fine-tuning? Traditional approaches face significant limitations: manual prompt engineering is labor-intensive with no guarantee of improvement, while SFT or RL requires substantial data and computational resources. To address this, the \textbf{Agent Practice} module integrates \textbf{Training-free Group Relative Policy Optimization (Training-free GRPO)}~\citep{youtu2025trainingfreegrpo}, enabling low-cost agent improvement through in-context experience accumulation without any gradient updates.

\begin{figure}[htbp]
  \begin{center}
    \includegraphics[width=0.99\linewidth]{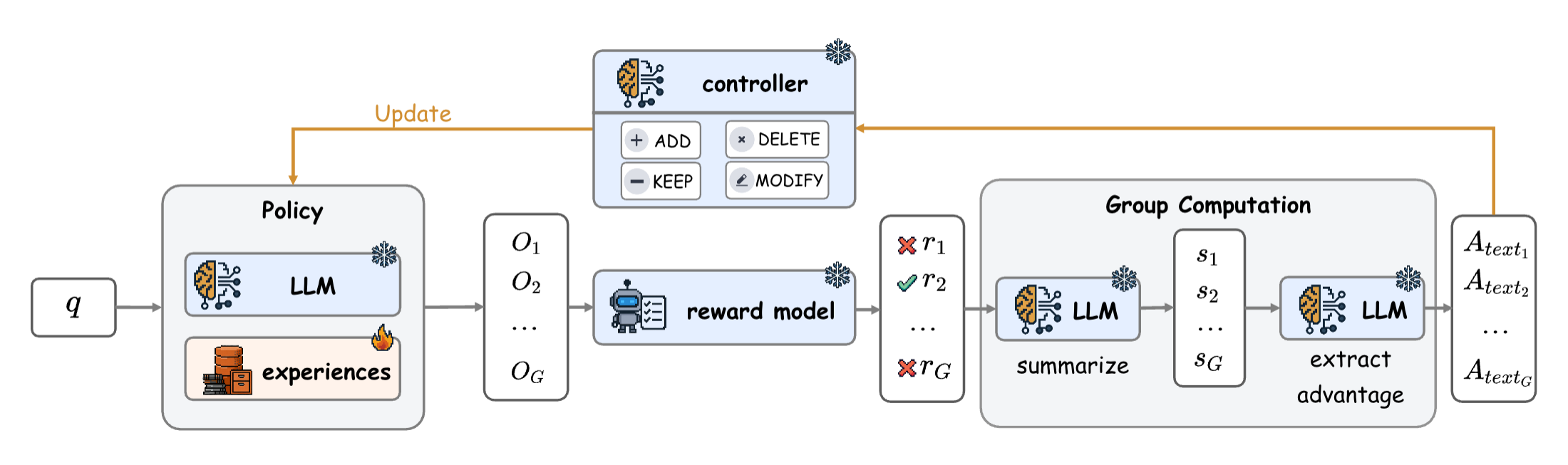}
  \end{center}
  \caption{Training-free GRPO mechanism. Given only a few dozen of samples, the agent performs multiple rollouts per task. An LLM evaluator assesses group relative trajectory quality, and distilling contextual memory of experiential knowledge by comparing successful and failed trials. During online testing, such learned experiences are injected as ``textual LoRA'' to guide reasoning.}
  \label{fig:tfgrpo}
\end{figure}

The core mechanism operates as follows: on a small training dataset (with or without ground truth), the agent performs multiple rollouts per task, generating diverse solution trajectories. Inspired by GRPO's group relative optimization, an evaluator assesses the relative quality of trajectories for the same task. 
Rather than calculating a numerical advantage in vanilla GRPO, the LLM introspects on these diverse outputs and distill a semantic group advantage, i.e., a textual learning direction derived from contrasting successful and failed trials. Such advantage optimizes the existing experiential knowledge, serving as a refined policy for subsequent epochs.
After completing practice, during online testing, these experiences are injected into the agent's context, acting as a form of ``textual LoRA'' that guides reasoning without modifying model weights.

Compared with traditional learning approaches, the agent practice module requires no gradient computation, works with API-based models where fine-tuning is inaccessible, leverages small training sets effectively, and allows experiences to be accumulated continuously.

\begin{figure}[htbp]
  \begin{center}
    \includegraphics[width=0.75\linewidth]{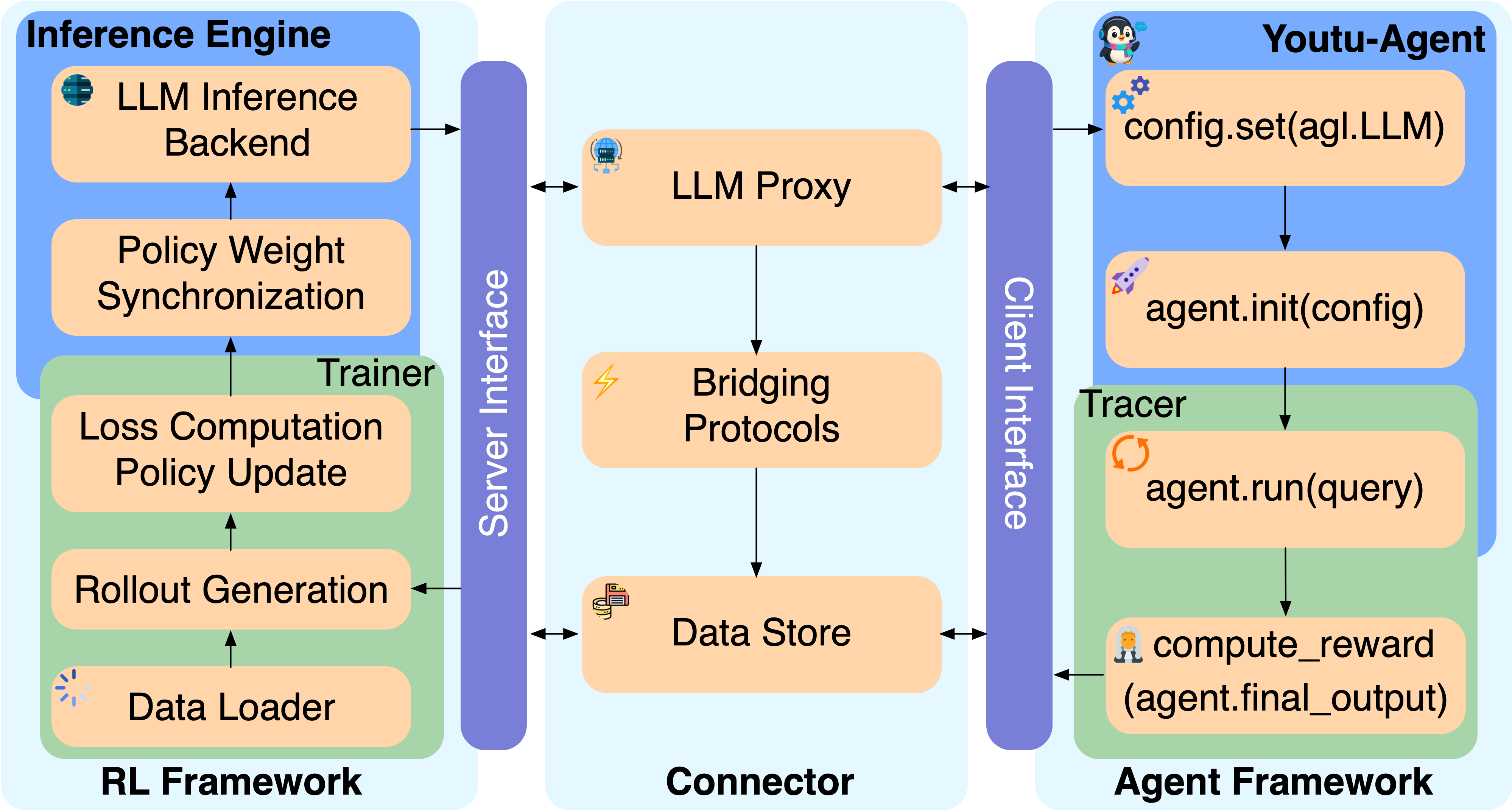}
  \end{center}
  \caption{An end-to-end RL training pipeline with Youtu-Agent. Left: Data flow in the RL training framework~\cite{sheng2025hybridflow}. Middle: Connector between RL framework and Agent framework~\cite{microsoft2025agentlightning}. Right: Data flow in the Youtu-Agent inference framework.}
  \label{fig:youtuagentrl}
\end{figure}

\subsection{Agent RL Training}
\label{sec:agentrl}

For applications requiring significant and lasting performance improvement, the \textbf{Agent RL} module provides a complete pipeline for end-to-end, production-grade reinforcement learning (see Figure~\ref{fig:youtuagentrl}).
We implement custom agent logic within our framework for inference,
and integrate with Agent-Lightning~\citep{microsoft2025agentlightning} as the connector between Youtu-Agent and modern RL framework such as VeRL~\cite{sheng2025hybridflow} for distributed training.
Such integration addresses two critical challenges that have historically limited agent RL: \textbf{scalability} to large distributed systems and \textbf{stability} over long-horizon tasks.

\textbf{Scalability Solutions.} Scaling agent RL presents unique challenges due to the complex, stateful nature of agent-environment interactions.
We address these through: (1) RESTful API wrapping that encapsulates agent execution environments as standardized services, enabling seamless distribution and load balancing; (2) Ray-based concurrency for highly parallel rollout collection across distributed workers; and (3) layered timeout logic with hierarchical controls at tool, step, and episode levels to handle various potential timeout issues without fatal failures.
Such infrastructure-level optimization enables stable scaling to 128 GPUs.

\textbf{Stability Solutions.} Long-horizon agent tasks exacerbate the ``entropy explosion'' problem where policies degenerate into repetitive or nonsensical actions. We mitigate this through: (1) filtering invalid and anomalous tool calls from training data to prevent learning degenerate patterns; 
(2) removing batch shuffling and reducing off-policy update iterations to prevent the policy from overfitting the outdated experience;
and (3) correcting bias of advantage estimation in turn-level GRPO training.
These techniques ensure stable training dynamics with consistent performance improvements across hundreds of training steps.

\section{Experiments}

We evaluate Youtu-Agent across four dimensions: general agent capabilities on standard benchmarks, effectiveness of automated generation mechanisms, low-cost improvement through the Agent Practice module, and scalable Agent RL training.

\subsection{Benchmark Performance}

To validate the effectiveness of our framework design and tool implementations, we evaluate Youtu-Agent on established agent benchmarks. Prioritizing accessibility and reproducibility, we build our agents entirely on open-source models---specifically the DeepSeek-V3 series---without relying on proprietary APIs such as Claude or GPT. We adopt a Plan-and-Execute paradigm, where a planner agent orchestrates task decomposition and scheduling while specialized executor agents handle subtasks.

\begin{figure}[htbp]
  \begin{center}
    \includegraphics[width=0.95\linewidth]{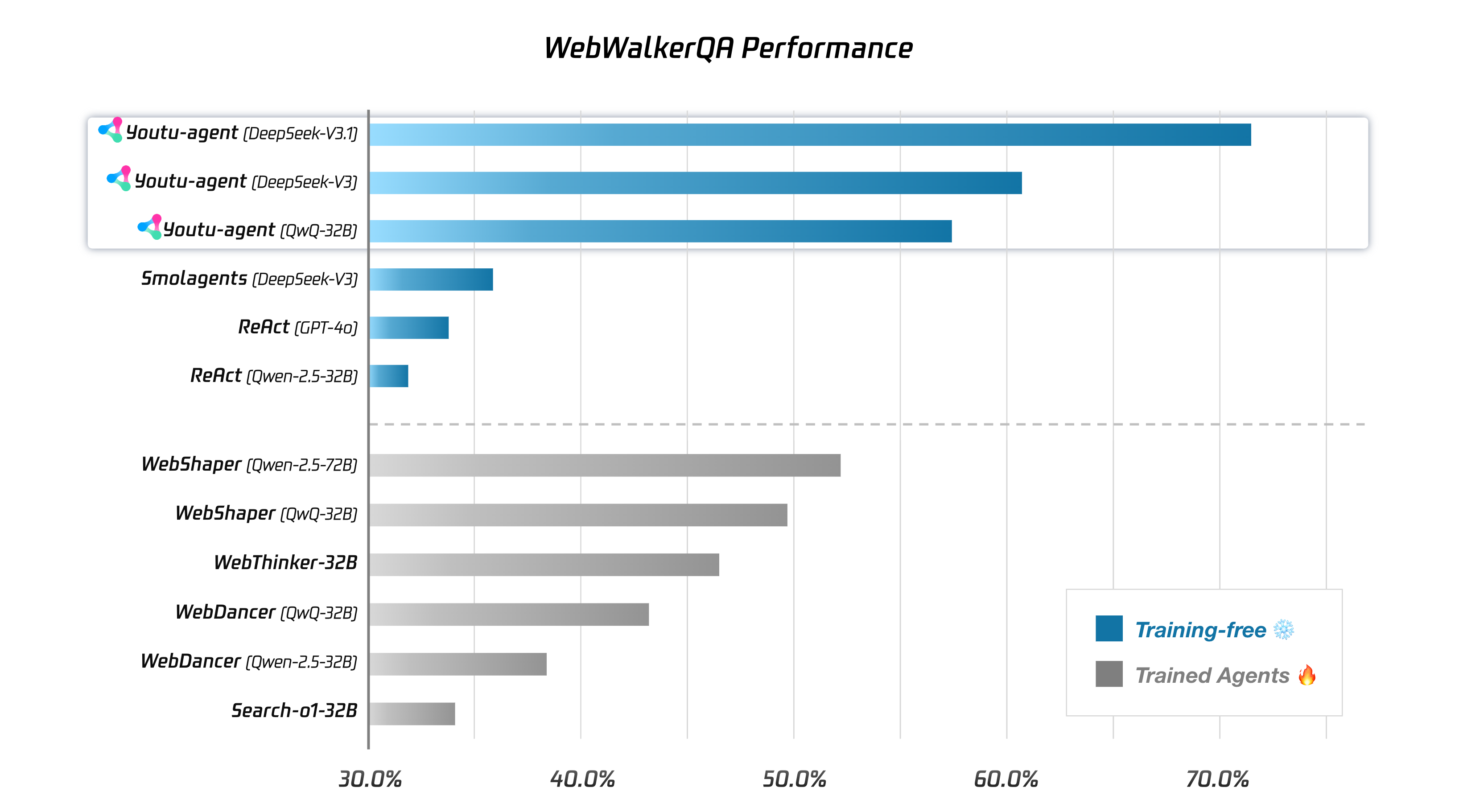}
  \end{center}
  \caption{Performance comparison on WebWalkerQA, including both training-free and trained agent approaches.}
  \label{fig:webwalkerqa}
\end{figure}

\textbf{WebWalkerQA.} This benchmark (680 questions) evaluates LLMs' ability to perform multi-step, deep web navigation and question answering over real websites. Equipped with web search and crawling tools, Youtu-Agent achieves \textbf{71.47\% pass@1 accuracy}, as shown in Figure~\ref{fig:webwalkerqa}. This result demonstrates robust environment interaction and multi-step exploration capabilities.

\textbf{GAIA (Text-only Subset).} GAIA~\citep{mialon2023gaia} (466 questions) tests real-world question answering requiring reasoning, multimodal understanding, web browsing, and tool-use proficiency. Since we use text-only models, we evaluate on the text-only subset. Beyond web tools, we provide document parsing and code execution capabilities to handle file attachments. Youtu-Agent achieves \textbf{72.8\% pass@1 accuracy}, demonstrating effective tool selection and deep reasoning abilities that validate our framework's design for general-purpose agent applications.

These results establish Youtu-Agent as a competitive open-source baseline without requiring proprietary APIs or closed-source systems.

\subsection{Automated Generation Effectiveness}

We evaluate the practical utility of Youtu-Agent's automated generation mechanisms through quantitative metrics and a qualitative case study.

\subsubsection{Evaluation Framework}

Given the absence of established benchmarks for automated agent generation, we construct \textbf{AgentGen-80}, a benchmark consisting of 80 diverse task descriptions ranging from simple information retrieval to complex multi-step automation. We assess generation quality across three dimensions:
\begin{itemize}
  \item \textbf{Configuration Validity (CV)}: Whether the generated YAML configuration is structurally valid and semantically complete.
  \item \textbf{Tool Executability (TE)}: Whether synthesized tools compile and execute successfully with expected functionality.
  \item \textbf{Task Completion (TC)}: End-to-end effectiveness---whether the generated agent successfully completes the specified task.
\end{itemize}

\subsubsection{Comparative Results}

We compare the Workflow mode and Meta-Agent mode on AgentGen-80. Table~\ref{tab:autogen} presents the results.

\begin{table}[htbp]
  \caption{Performance comparison of Workflow and Meta-Agent generation modes on AgentGen-80.}
  \label{tab:autogen}
  \begin{center}
    \begin{tabular}{lccc}
      \toprule
      \textbf{Generation Mode} & \textbf{CV} & \textbf{TE} & \textbf{TC} \\
      \midrule
      Workflow Mode            & 100\%       & 81.25\%     & 65.00\%     \\
      Meta-Agent Mode          & 98.75\%     & 82.50\%     & 68.75\%     \\
      \bottomrule
    \end{tabular}
  \end{center}
\end{table}

The Workflow mode achieves \textbf{100\% configuration validity}, demonstrating the robustness of its deterministic four-stage pipeline. The Meta-Agent mode achieves 98.75\% validity---the occasional failures occur when the final \texttt{create\_agent\_config} tool call fails to produce well-formed output. For \textbf{Tool Executability (TE)}, both modes use the same tool synthesis module, achieving 81.25\% and 82.50\% respectively with no significant difference. For \textbf{Task Completion (TC)}, we manually test the generated agents through dialogue to verify expected behavior, yielding 65.00\% and 68.75\% completion rates. The Meta-Agent mode shows slightly better end-to-end effectiveness.

\subsubsection{Case Study}

We illustrate the Meta-Agent mode with a concrete example. For the user query \textit{``Summarize today's trending papers on multi-agent system and download PDFs,''} the Architect Agent first calls \texttt{search\_tool} to search the existing library, finding the \texttt{arxiv} toolkit with paper download capabilities. Finding no existing tool for fetching daily paper updates, it calls \texttt{create\_tool} to synthesize a new tool \texttt{fetch\_daily\_papers}. Finally, it calls \texttt{create\_agent\_config} to assemble the configuration file shown below:

\begin{lstlisting}[style=yaml]
agent:
  name: Papers_Analyzer_Agent
  instructions: |
    You are a specialized research assistant focused on analyzing
    daily papers with expertise in agent technologies.
    ...
toolkits:
  search: {"activated_tools": ["search", "web_qa"]}  # retrieved from library
  arxiv: {"activated_tools": ["download_papers"]}    # retrieved from library
  fetch_daily_papers: {}                             # synthesized tool
\end{lstlisting}

Within the \texttt{create\_tool} invocation, a subagent queries the web to retrieve related API documentation, then implements the tool in Python. The core implementation is shown below, and the synthesized tool is exposed via MCP (Model Context Protocol) to the generated agent, enabling seamless integration with the execution loop.

\begin{lstlisting}[style=yaml]
def fetch_daily_papers(date: str) -> str:
    """Crawl daily papers from aggregation site. Return paper infos in str.

    Args:
        date (str): date in format YYYY-MM-DD
    """
    papers = list_daily_papers(date=date)
    return "\n".join([f"{asdict(paper)}" for paper in papers])
\end{lstlisting}

\subsection{Agent Practice Module Results}

As introduced in Section~\ref{sec:practice}, the Agent Practice module integrates Training-free GRPO to enable low-cost agent improvement through experience accumulation without parameter updates. We validate its effectiveness on challenging mathematical reasoning benchmarks AIME 2024 and AIME 2025~\citep{AIME}, using Mean@32 for robust measurement.

\textbf{Experimental Setup.}
Our evaluation is based on ReAct~\citep{yao2022react} with a code interpreter tool.
During the learning stage, we randomly sample 100 problems from the DAPO-Math-17K~\citep{dapo} dataset and run 3 epochs with group size $5$ and temperature $0.7$. For online testing, the temperature is set to $0.3$.

\begin{table}[htbp]
  \caption{Performance comparison on AIME 2024 and 2025 (Mean@32). Training-free GRPO achieves +2.7\% and +5.4\% absolute improvements over the ReAct baseline with only 100 training examples and zero gradient updates.}
  \label{tab:tfgrpo}
\centering
\begin{tabular}{lclcc}
\toprule
\textbf{Method} & \textbf{Learning Cost} & \textbf{Model} & \textbf{AIME24} & \textbf{AIME25} \\
\midrule
ReAct~\citep{yao2022react} & - & \textit{Qwen2.5-32B-Instruct} & 29.6 & 23.1 \\
ZeroTIR~\citep{zerotir} & $\approx\$20,000$ & \textit{Qwen2.5-32B-Instruct}& 56.7 &  33.3 \\
SimpleTIR~\citep{simpletir} & $\approx\$20,000$ & \textit{Qwen2.5-32B-Instruct}& 59.9 & 49.2 \\
ReTool~\citep{retool} & $\approx\$10,000$ & \textit{Qwen2.5-32B-Instruct}& 67.0 & 49.3 \\
AFM~\citep{oppo2025chainofagents} & $\approx\$10,000$ & \textit{Qwen2.5-32B-Instruct}& 66.7 & 59.8 \\
\midrule
ReAct~\cite{yao2022react} & - & \textit{DeepSeek-V3.1-Terminus} & 80.0 & 67.9 \\
+ Training-Free GRPO {\footnotesize(w/o ground truths)} & $\approx\$18$ & \textit{DeepSeek-V3.1-Terminus} & 80.7 & 68.9 \\
+ Training-Free GRPO {\footnotesize(w/ ground truths)}& $\approx\$18$ & \textit{DeepSeek-V3.1-Terminus} & 82.7 & 73.3 \\
\bottomrule
\end{tabular}
\end{table}

\begin{figure}[htbp]
  \begin{center}
    \includegraphics[width=0.85\linewidth]{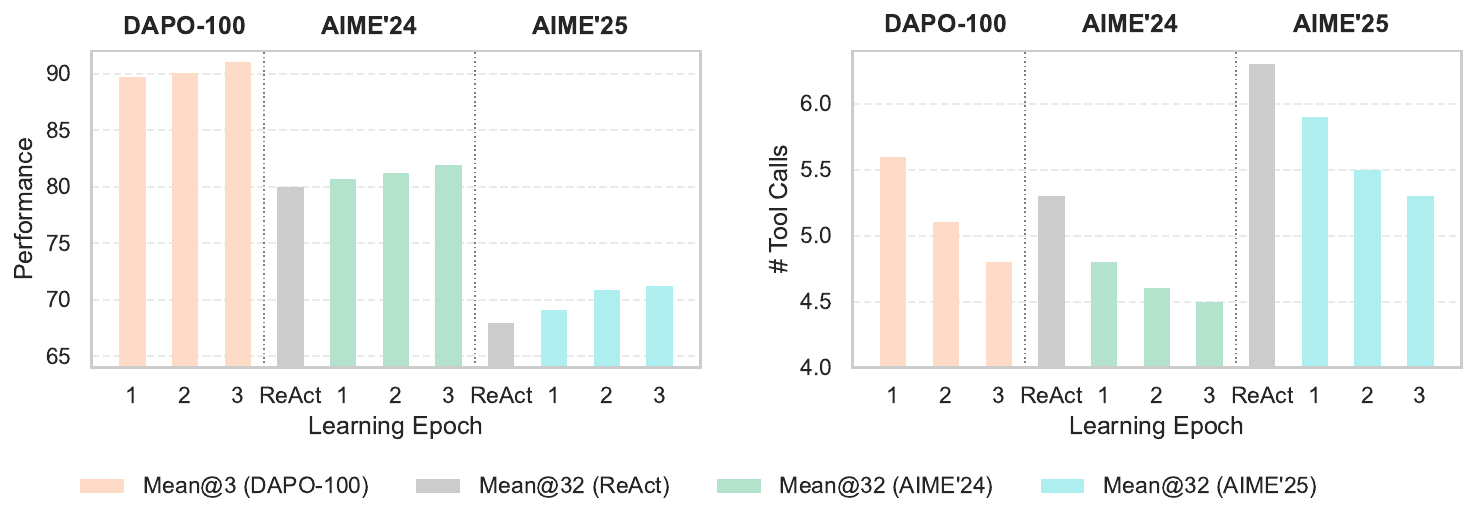}
  \end{center}
  \caption{During the learning stage of Training-free GRPO, performance improves steadily while tool usage becomes more efficient.}
  \label{fig:tfgrpo_dynamics}
\end{figure}

As shown in Table~\ref{tab:tfgrpo}, when ground truths are available for the training samples, Training-free GRPO achieves +2.7\% on AIME 2024 and +5.4\% on AIME 2025 compared to ReAct, demonstrating the effectiveness of learned experiences. Notably, the variant without ground truths still achieves strong results, validating applicability to domains where labeled data is scarce.
Compared to RL methods requiring 10,000+ training samples and approximately \$10,000 for fine-tuning a 32B model, Training-free GRPO achieves better performance with only 100 samples and approximately \$18 learning costs, offering a highly effective and accessible pathway for real-world applications.
Furthermore, as shown in Figure~\ref{fig:tfgrpo_dynamics}, both learning and validation performance improve steadily across epochs, and the average number of tool calls is decreasing, which indicates that the agent learns more efficient problem-solving strategies.

\subsection{Agent RL Training Results}

As introduced in Section~\ref{sec:agentrl}, our Agent RL module proposes modifications to both infrastructure and algorithm implementation of Agent-Lightning~\cite{microsoft2025agentlightning} to enable scalable and stable end-to-end RL training.
The effectiveness of our optimization is confirmed through both training efficiency and stability on multiple benchmarks.

\begin{figure}[htbp]
  \centering
  \begin{subfigure}[b]{0.4\linewidth}
    \includegraphics[width=\linewidth]{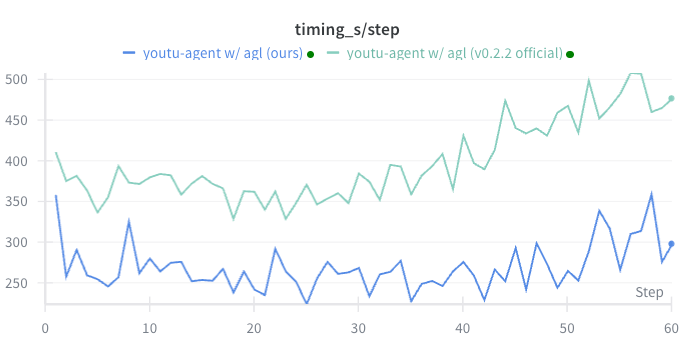}
    \caption{Time of an iteration step.}
  \end{subfigure}
  \hspace{1em}
  \begin{subfigure}[b]{0.4\linewidth}
    \includegraphics[width=\linewidth]{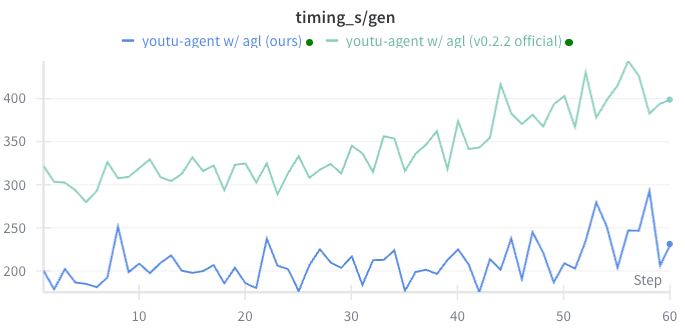}
    \caption{Time of the rollout generation part.}
  \end{subfigure}
  \caption{The comparison of our Agent RL module with the official Agent-Lightning in training efficiency.}
  \label{fig:rl_efficiency}
\end{figure}

\textbf{Efficiency.} 
Compared with the Agent-Lightning (v0.2.2 official version),
our infrastructure optimizations such as concurrency control, RESTful API wrapping, and hierarchical timeout logic reduce iteration time by approximately \textbf{40\%}, enabling efficient scaling to 128 GPUs without timeout issues (see Figure~\ref{fig:rl_efficiency}).

\textbf{Effectiveness.} To ensure fair comparison,
we perform RL training of Qwen2.5-7B-Instruct on two typical tasks: 1) math/code for reasoning capabilities,
and 2) search for information-seeking capabilities.

For the math/code task, we follow the experimental settings of ReTool~\cite{retool} where a code interpreter (sandbox) is prepared as an external tool for assistance of solving mathematic problems.
As shown in Figure~\ref{tab:agentrl_maths},
evaluation results confirm that the accuracy gains of 35\% and 22\% are respectively achieved on AIME24 and AIME25.

\begin{table}[htbp]
  \caption{Results on Maths benchmarks (step 500). Qwen2.5-7B shows consistent improvements on AIME24 and AIME25 via our end-to-end RL module with Youtu-Agent.}
  \label{tab:agentrl_maths}
  \begin{center}
    \begin{tabular}{lccc}
      \toprule
      \textbf{Dataset} & \textbf{Before} & \textbf{After} & \textbf{$\Delta$} \\
      \midrule
      AIME24         &   0.10   &   0.45   &  +0.35    \\
      AIME25            &  0.09   &   0.31    &  +0.22     \\
      \bottomrule
    \end{tabular}
  \end{center}
\end{table}

For the search task,
we follow the experimental settings of SearchR1~\cite{jin2025search} where a local wikipedia-2018 retrieval service (E5-HNSW64 embeddings) is prepared as an external tool for solving general QA and multi-hop QA problems.
As shown in Table~\ref{tab:agentrl_search},
consistent improvements are observed across all benchmarks with accuracy gains of 17\%, 19\%, 21\%, 8\%, 17\%, 13\%, and 10\% respectively on TriviaQA, PopQA, NatureQuestions, MusiQue, HotpotQA, Bamboogle, and 2WikiMultiHop.

\begin{table}[htbp]
  \caption{Results on general QA and multi-hop QA benchmarks (step 200). Qwen2.5-7B-Instruct shows consistent improvements across all benchmarks via our end-to-end RL module with Youtu-Agent.}
  \label{tab:agentrl_search}
  \begin{center}
    \begin{tabular}{lccc}
      \toprule
      \textbf{Dataset} & \textbf{Before} & \textbf{After} & \textbf{$\Delta$} \\
      \midrule
      TriviaQA         & 0.37            & 0.54           & +0.17             \\
      PopQA            & 0.16            & 0.35           & +0.19             \\
      NaturalQuestions & 0.24            & 0.45           & +0.21             \\
      MuSiQue          & 0.06            & 0.14           & +0.08             \\
      HotpotQA         & 0.21            & 0.38           & +0.17             \\
      Bamboogle        & 0.23            & 0.36           & +0.13             \\
      2WikiMultiHop    & 0.22            & 0.32           & +0.10             \\
      \bottomrule
    \end{tabular}
  \end{center}
\end{table}

\begin{figure}[htbp]
  \centering
  \begin{subfigure}[b]{0.48\linewidth}
    \includegraphics[width=\linewidth]{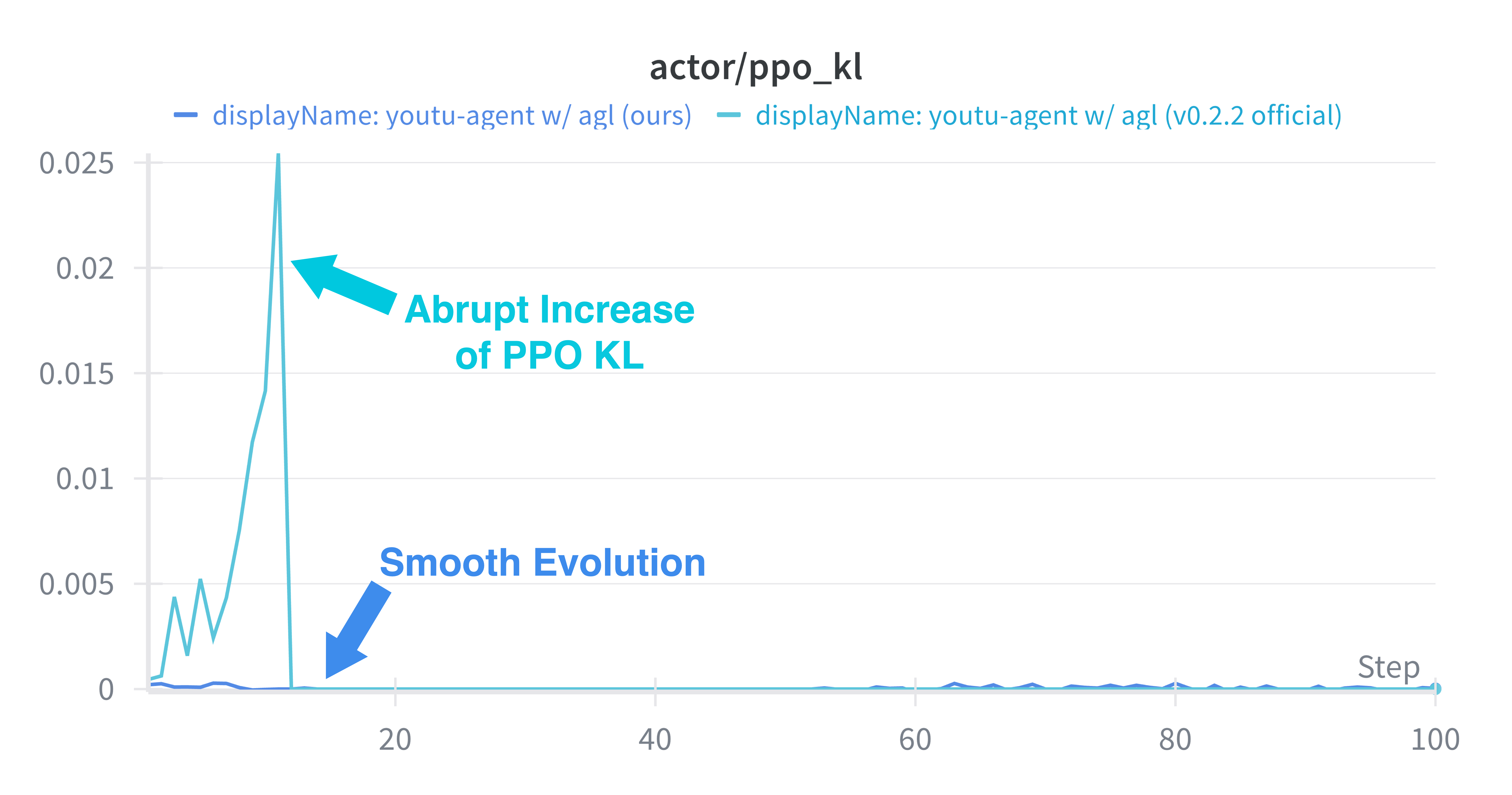}
    \caption{PPO KL Divergence}
  \end{subfigure}
  \begin{subfigure}[b]{0.48\linewidth}
    \includegraphics[width=\linewidth]{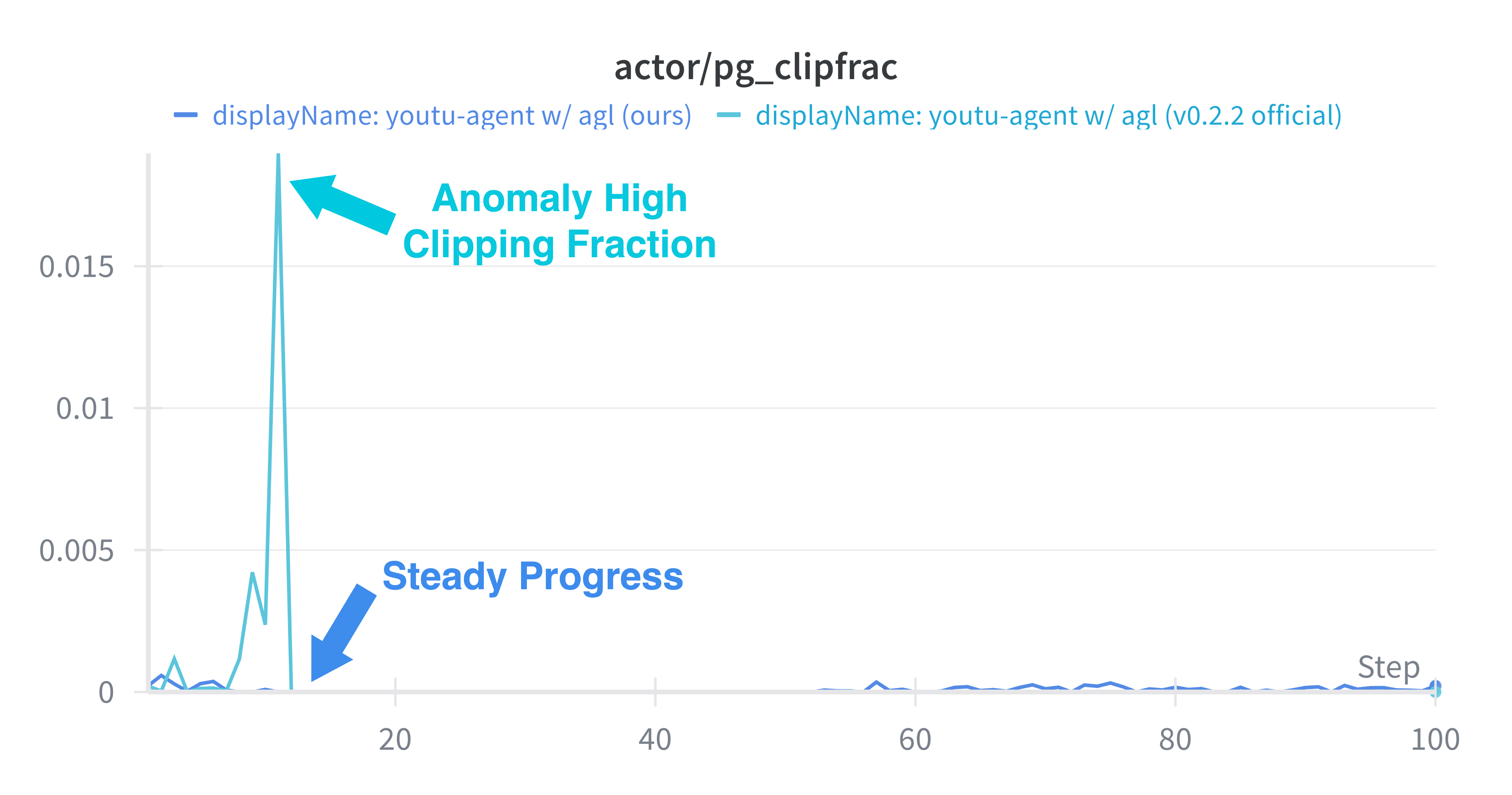}
    \caption{Policy Gradient Clip Fraction}
  \end{subfigure}
  \vspace{0.3em}
  \begin{subfigure}[b]{0.48\linewidth}
    \includegraphics[width=\linewidth]{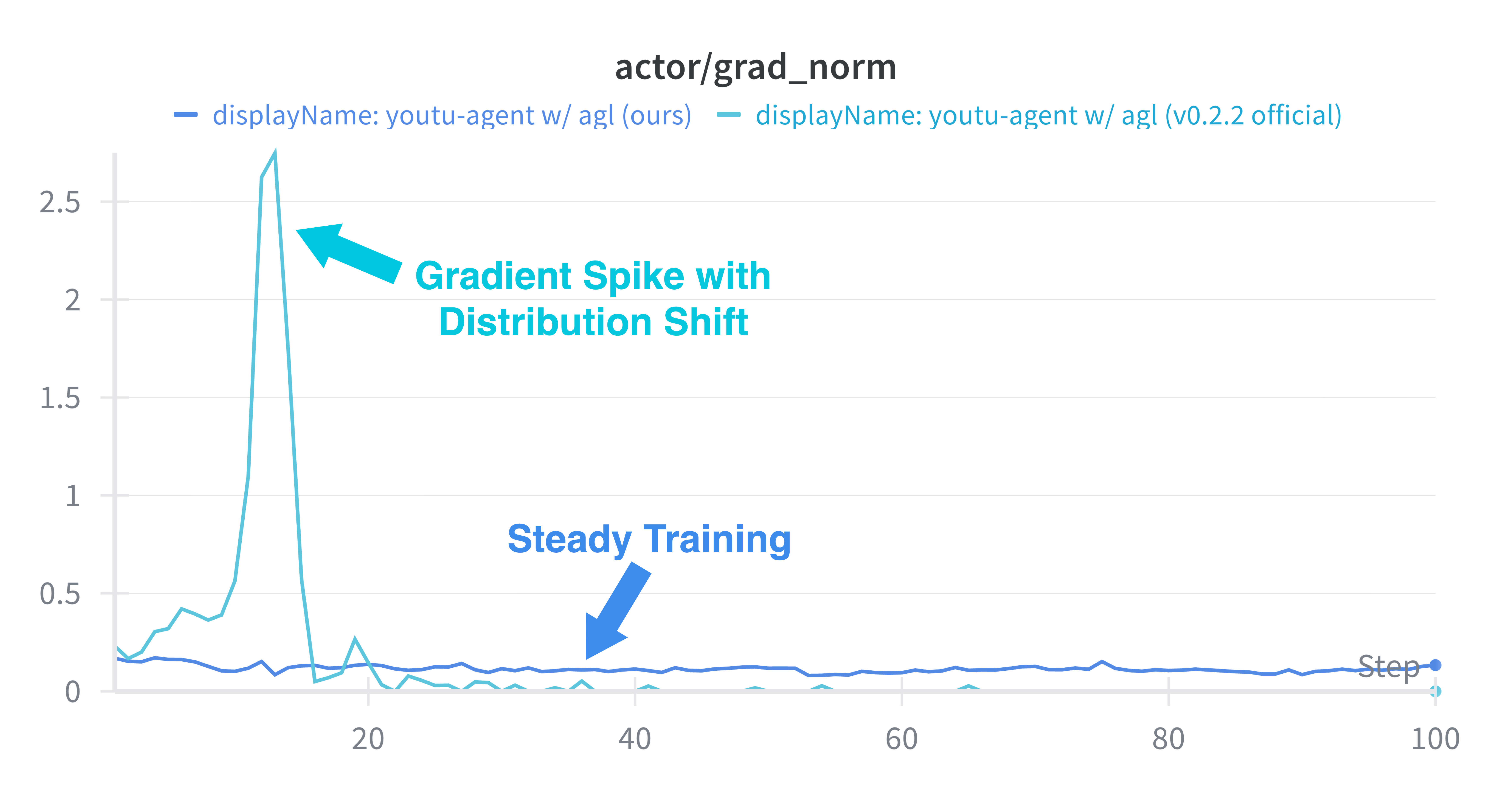}
    \caption{Gradient Norm}
  \end{subfigure}
  \begin{subfigure}[b]{0.48\linewidth}
    \includegraphics[width=\linewidth]{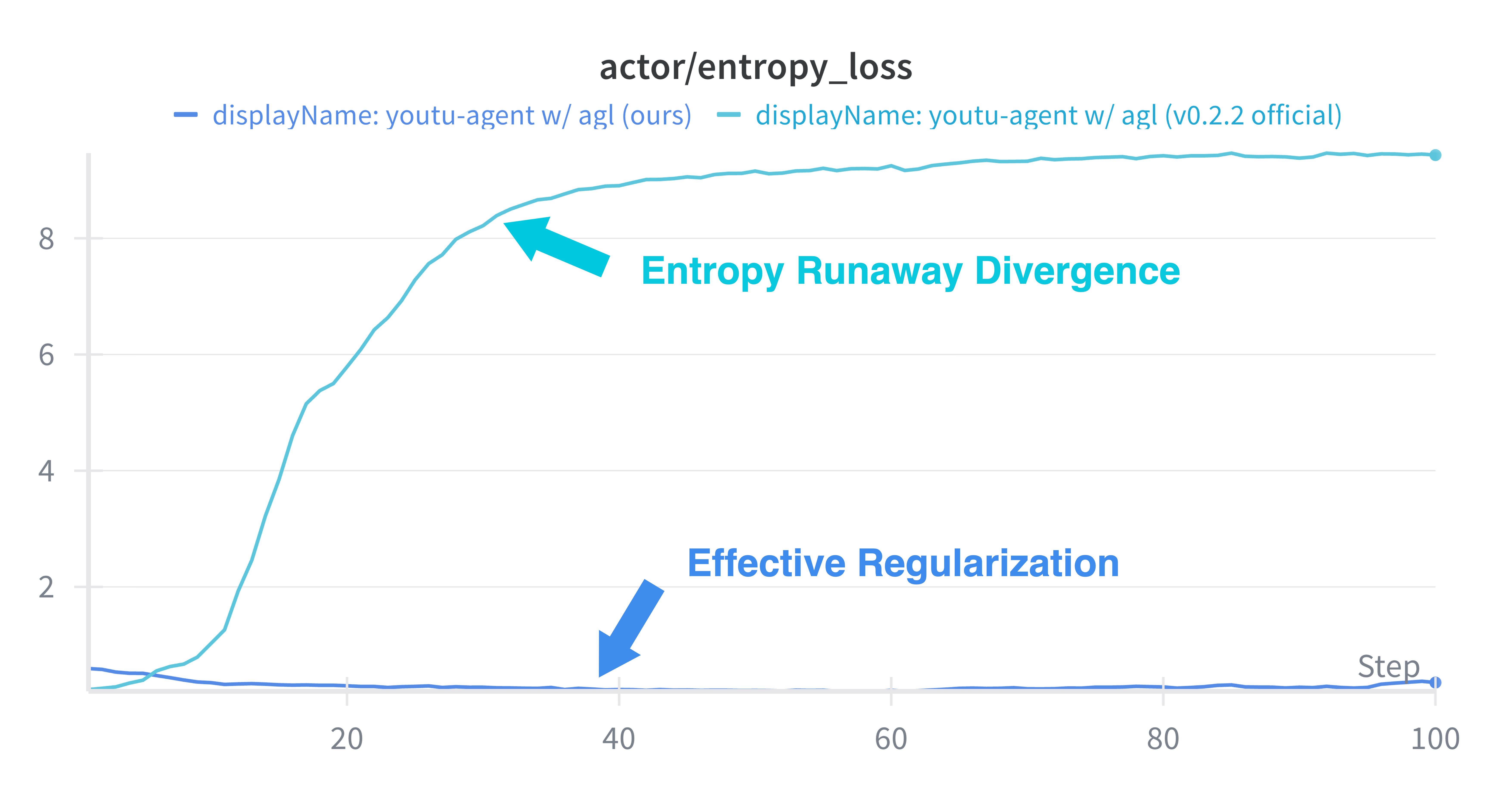}
    \caption{Entropy Loss}
  \end{subfigure}
  \vspace{0.3em}
  \begin{subfigure}[b]{0.48\linewidth}
    \includegraphics[width=\linewidth]{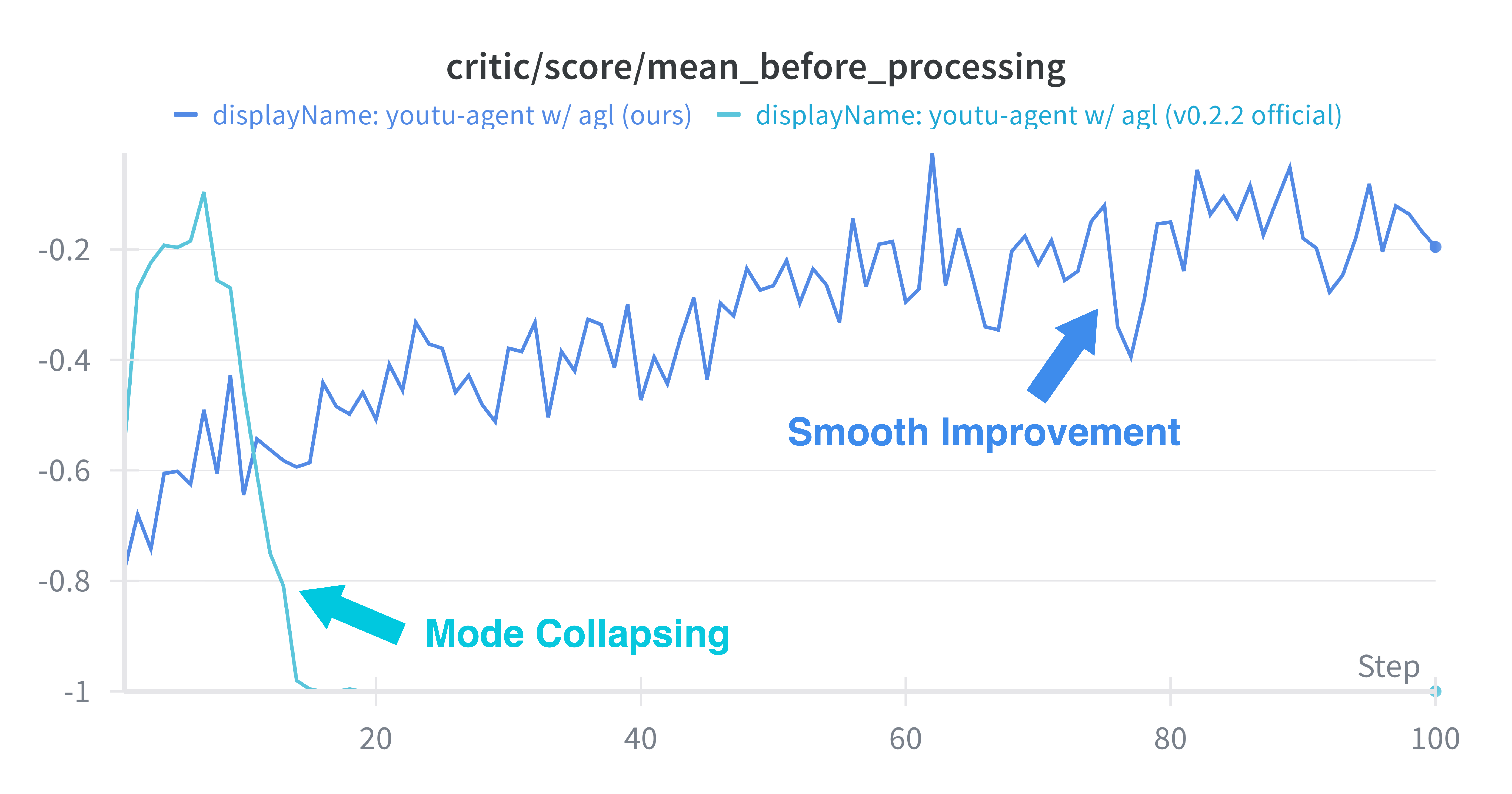}
    \caption{Critic Score}
  \end{subfigure}
  \begin{subfigure}[b]{0.48\linewidth}
    \includegraphics[width=\linewidth]{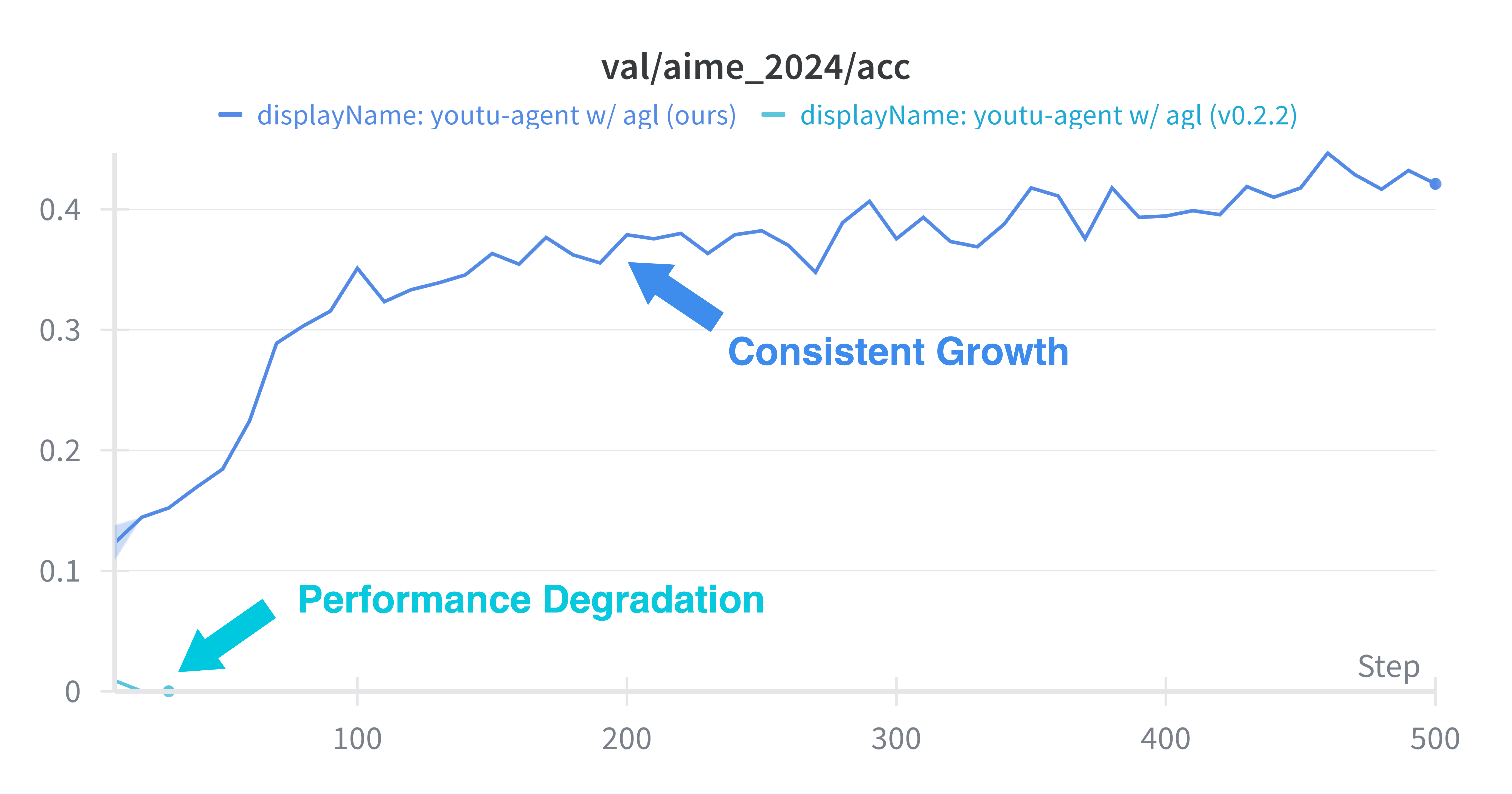}
    \caption{Validation Accuracy}
  \end{subfigure}
  \caption{Training dynamics on ReTool dataset. Comparison between our optimized agent RL module and their official v0.2.2 release. Our improvements ensure stable KL divergence, controlled gradient norm, steady critic score growth, and consistent validation accuracy increase.}
  \label{fig:rl_dynamics}
\end{figure}

\textbf{Training Dynamics.} Figure~\ref{fig:rl_dynamics} illustrates the training dynamics for comparison with the official Agent-Lightning release.
After our improvements addressing ``entropy explosion'',
the actor's KL divergence and gradient norm remain stable.
It can be observed that critic score increases steadily,
which is in line with the validation accuracy.

These experiments demonstrate that Youtu-Agent not only provides a robust execution framework but also supports effective model training for building continuously improving agents.

\section{Related Work}

\subsection{Agent Frameworks}

Recent years have seen a proliferation of agent frameworks. \textbf{MetaGPT}~\citep{hong2023metagpt}, \textbf{AutoGen}~\citep{wu2024autogen}, and \textbf{ChatDev}~\citep{qian2024chatdev} have popularized multi-agent collaboration, assigning static roles to LLMs to solve complex tasks. While effective, these frameworks often rely on predefined roles and heavy prompt tuning. Youtu-Agent advances this paradigm by automating the generation of agent configurations themselves, including executable tool code and optimized prompts, rather than just orchestrating static roles.

\subsection{Automated Agent Design}

Research into automated agent design, such as \textbf{ADAS}~\citep{hu2024adas} and \textbf{AutoAgents}~\citep{chen2023autoagents}, has explored generating agents from high-level descriptions. Youtu-Agent distinguishes itself by not only generating the agent's design but also synthesizing the necessary tool code (Python) and providing a unified pipeline that extends from automated generation to continuous optimization.

\subsection{Agent Optimization and Learning}

Optimizing agents typically involves either prompt optimization or parameter updates. \textbf{Reflexion}~\citep{shinn2023reflexion} and \textbf{ReAct}~\citep{yao2022react} use verbal reinforcement and reasoning traces to improve performance without training. On the training side, \textbf{Toolformer}~\citep{schick2023toolformer} and \textbf{Gorilla}~\citep{patil2024gorilla} focus on fine-tuning for tool use. Youtu-Agent integrates both approaches: the \textbf{Practice} module enables low-cost, inference-time improvement through experience accumulation, and the \textbf{Agent RL} module provides a full-scale RL pipeline for end-to-end parameter optimization, addressing the stability issues often encountered in PPO/RLHF for agents.

\section{Conclusion}

We presented Youtu-Agent, a modular framework for automatic agent generation and continuous improvement. The framework addresses two fundamental challenges in LLM-based agents: high configuration costs and static capabilities.

\textbf{Key Contributions.} (1) A modular, YAML-based architecture that decouples environments, tools, and context management for flexible agent composition. (2) Dual-paradigm automated generation through workflow pipelines and meta-agent interaction. (3) An Agent Practice module for low-cost optimization using small datasets without gradient updates. (4) An Agent RL module for scalable and stable end-to-end reinforcement learning training.

\textbf{Empirical Results.} Youtu-Agent achieves 71.47\% on WebWalkerQA and 72.8\% on GAIA using only open-source models. Our automated generation achieves over 81\% tool executability and up to 68.75\% task completion. The Practice module improves AIME performance by +4.8\%/+7.0\% with only 100 samples and \$18 cost. The Agent RL module enables 40\% speedup and stable 128-GPU scaling, improving Qwen2.5-7B from 10\% to 45\% accuracy on AIME 2024.

\textbf{Future Directions.} We plan to expand the framework with more environment integrations, enhanced multi-agent collaboration capabilities, and more sophisticated experience accumulation strategies.

\section{Application}

To make youtu-agent more accessible to end users, we present \textbf{Tip}, an on-device, multimodal desktop assistant that keeps data local, integrates Youtu-Agent, and can directly drive a GUI agent. Key functions include:

\begin{itemize}
\item Youtu-Agent inside: load and run existing Youtu-Agent configs to
handle bash commands, file management or more complex tasks within a unified
UI.
\item Proactive intent and context completion: automatically capture
relevant screen/text context and surface intents, no manual copy/paste or
typing needed.
\item GUI Agent with skills: automate desktop actions end to end, then save
and replay “GUI skills” as reusable workflows for continual improvement.
\item On-device model support: run local models to keep data private and
secure.
\end{itemize}

Tip let users ask at their fingertip without further tiresome operation while achieving complex goal smoothly, locally and safely. We aim for this project to catalyze the development and practical adoption of the Agent technology community in future.

\begin{figure}[htbp]
  \begin{center}
    \includegraphics[width=\linewidth]{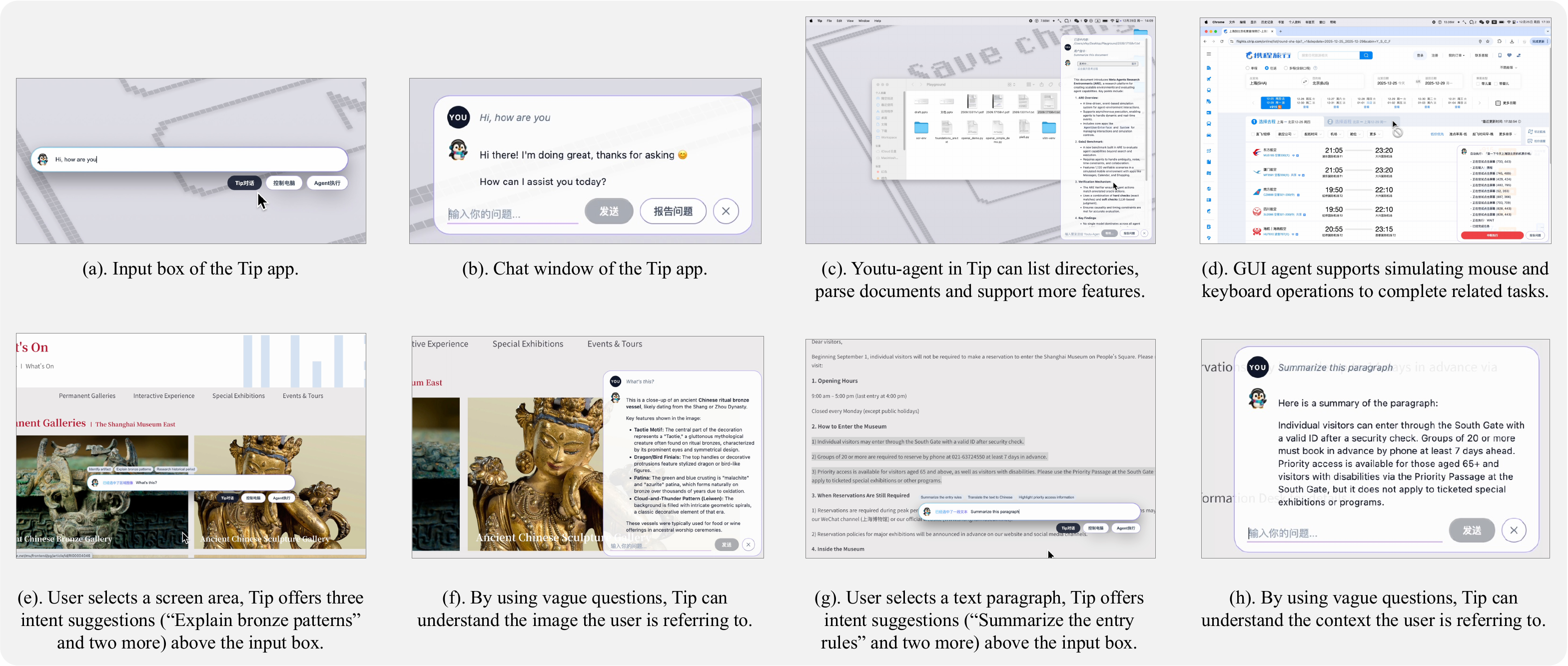}
  \end{center}
  \caption{Demonstrations of the Tip application.}
  \label{fig:tip_demo}
\end{figure}

\section*{Contributions}

\textbf{Authors} \quad
Yuchen Shi\textsuperscript{\rm 1*}\quad Yuzheng Cai\textsuperscript{\rm 1,2*\textdagger} \quad Siqi Cai\textsuperscript{\rm 1*}\quad Zihan Xu\textsuperscript{\rm 1*} \quad Lichao Chen\textsuperscript{\rm 1,3\textdagger} \quad Yulei Qin\textsuperscript{\rm 1} \quad Zhijian Zhou\textsuperscript{\rm 1,2} \quad Xiang Fei\textsuperscript{\rm 1}\quad Chaofan Qiu\textsuperscript{\rm 1}\quad Xiaoyu Tan\textsuperscript{\rm 1}\quad Gang Li\textsuperscript{\rm 1} \quad Zongyi Li\textsuperscript{\rm 1} \quad Haojia Lin\textsuperscript{\rm 1} \quad Guocan Cai\textsuperscript{\rm 1}\quad Yong Mao\textsuperscript{\rm 1} \quad Yunsheng Wu\textsuperscript{\rm 1} \quad Ke Li\textsuperscript{\rm 1\(\spadesuit\)}\quad Xing Sun\textsuperscript{\rm 1\faEnvelopeO}

\textbf{Affiliations} \quad
\textsuperscript{\rm 1}Tencent Youtu Lab\quad \textsuperscript{\rm 2}Fudan University\quad \textsuperscript{\rm 3}Xiamen University

* Equal Contribution\quad \textdagger Work during Internship at Tencent \quad\(\spadesuit\) Project Lead\quad \faEnvelopeO Correspondence: winfredsun@tencent.com

\paragraph{Acknowledgments}
We thank the teams behind Agent-Lightning for their open-source contributions that enabled our scalable RL training experiments.

\setcitestyle{numbers,square}
\bibliography{main}

\end{document}